# Preoperative Prediction of Microvascular Invasion in Hepatocellular Carcinoma by Integrating Multimodal Ultrasound and Clinical Data: A Multicenter Study

Jun Cheng[1,2], Yuanyuan Kong[1,2], Qing Huang[1,2], Xiaotong Tan[1,2], Licong Dong[3], Yulong Han[1,2], Wufeng Xue[1,2], Ruobing Huang[1,2], Dong Ni[1,2], Qi Yang[3]*, Jie Yu[4]*, Ping Liang[4]*

1. National-Regional Key Technology Engineering Laboratory for Medical Ultrasound, Guangdong Key Laboratory for Biomedical Measurements and Ultrasound Imaging, School of Biomedical Engineering, Shenzhen University Medical School, Shenzhen University, Shenzhen, China

2. Marshall Laboratory of Biomedical Engineering, Shenzhen University, Shenzhen, China

3. Department of Ultrasound, Peking University Shenzhen Hospital, Shenzhen, China

4. Department of Interventional Ultrasound, Senior Department of Oncology, Chinese PLA General Hospital, Beijing, China

* Corresponding authors

## Abstract

**Background:** Microvascular invasion (MVI) predicts recurrence and survival in hepatocellular carcinoma (HCC) but requires postoperative histopathology for diagnosis. We developed and validated a model integrating multimodal ultrasound and clinical data for preoperative MVI prediction.

**Methods:** This multicenter study included 489 patients with HCC from eight centers. All patients had B-mode ultrasound (BUS), color Doppler flow imaging (CDFI), dynamic contrast-enhanced ultrasound (DCE-US), and clinical information. Data from seven centers (n = 421) were used for model development with five-fold cross-validation; data from the remaining center (n = 68) formed an independent external validation cohort. The proposed multimodal information fusion network used modality-specific encoders, a hemodynamic temporal change module for bidirectional DCE-US perfusion changes, and a representation consistency learning module to align heterogeneous ultrasound representations before Transformer-based fusion.

**Results:** In external validation, DCE-US achieved the highest single-modality area under the receiver operating characteristic curve (AUC; 0.8545 ± 0.0198), versus clinical information (0.6715 ± 0.0156), CDFI (0.6435 ± 0.0344), and BUS (0.6087 ± 0.0417). Pixel-difference sampling and the proposed temporal module outperformed alternative sampling and video representation methods. The full model achieved the best performance, with an AUC of 0.8953 ± 0.0180, accuracy of 81.18% ± 2.83%, sensitivity of 86.40% ± 6.69%, and specificity of 78.14% ± 6.28.

**Conclusions:** Integrating multimodal ultrasound and clinical information enabled promising preoperative MVI prediction in HCC. DCE-US was the main source of predictive information, while BUS, CDFI, and clinical information provided complementary value. The proposed framework may support preoperative risk stratification and individualized clinical decision-making.

## 1. Background

Hepatocellular carcinoma (HCC) accounts for approximately 90% of primary liver cancers and remains a leading cause of cancer-related mortality worldwide [1,2]. Although liver resection, transplantation, and local ablation offer potentially curative treatment, postoperative recurrence remains common, even after R0 resection [3]. Microvascular invasion (MVI), pathologically defined as microscopic tumor cell nests within endothelium-lined vascular spaces near a tumor, is an important indicator of aggressive tumor behavior and is strongly associated with early recurrence and poor survival after hepatectomy [4,5]. The type of involved vessels and the distance of invasion from the tumor margin may further stratify recurrence and mortality risks [6,7]. Despite its strong prognostic significance, MVI can currently be confirmed only by histopathological examination of resected specimens. Accurate preoperative prediction of MVI could therefore facilitate surgical planning, perioperative management, and individualized postoperative surveillance [8].

Ultrasound is widely used for HCC evaluation because it enables real-time assessment of tumor morphology, vascularity, and perfusion [9]. A systematic review reported a pooled area under the receiver operating characteristic curve (AUC) of approximately 0.81 for ultrasound radiomics models in MVI prediction, indicating that ultrasound contains quantitatively measurable information associated with MVI [10]. Different ultrasound modalities characterize distinct aspects of tumor biology. B-mode ultrasound (BUS) depicts lesion morphology, margins, internal echogenicity, and peritumoral structures; color Doppler flow imaging (CDFI) visualizes macroscopic intratumoral and peritumoral blood flow; and dynamic contrast-enhanced ultrasound (DCE-US) captures the temporal evolution of enhancement and washout, thereby reflecting tumor microcirculatory perfusion [11]. Integrating these complementary sources of information may provide a more comprehensive representation of MVI-related phenotypes than any single modality. However, effectively integrating these heterogeneous ultrasound modalities together with clinical information remains methodologically challenging.

Radiomics and deep learning have increasingly been applied to preoperative MVI prediction using CT, MRI, PET/CT, and ultrasound [12–14]. In ultrasound-based studies, a DCE-US radiomics nomogram achieved an AUC of 0.788, while a model combining CEUS LI-RADS with clinical features achieved an AUC of 0.840 [15,16]. Deep learning models using DCE-US images or videos subsequently achieved AUCs ranging from 0.835 to 0.865 [17–19]. However, most of these studies focused primarily on DCE-US alone and did not explicitly model the complementary relationships among grayscale morphology, macroscopic blood flow, and dynamic microvascular perfusion. Furthermore, a cross-institutional study reported external-validation AUCs of only 0.667–0.688 despite stronger internal validation performance, highlighting the challenge of model generalizability across centers [20]. A recent multicenter study integrating conventional ultrasound, Sonazoid-enhanced ultrasound, and biochemical indicators reported relatively strong predictive performance; however, its model relied primarily on manually assessed imaging features and logistic regression [21]. Collectively, these studies demonstrate the potential of ultrasound-based MVI prediction while underscoring the need for a unified deep learning framework that can learn representations from heterogeneous ultrasound data, explicitly capture dynamic perfusion changes, and effectively integrate complementary information across modalities.

To address these gaps, we developed a multimodal information fusion neural network (MIFNN) for preoperative prediction of MVI in HCC. The proposed framework has three principal features. First, it integrates BUS, CDFI, DCE-US sequences, and clinical information, jointly characterizing tumor morphology, macroscopic blood flow, dynamic perfusion, and clinical background. Second, a hemodynamic temporal change module explicitly models bidirectional inter-frame perfusion changes in DCE-US, while a representation consistency learning module aligns heterogeneous ultrasound representations before Transformer-based fusion. Third, the model was developed using data from seven medical centers and evaluated in an independent cohort from an eighth center to assess cross-center generalizability.

## 2. Methods

### 2.1 Study design

This study was approved by the ethics committees of all participating centers and registered on ClinicalTrials.gov (NCT03871140). All patients provided written informed consent. Personal health information was de-identified.

The overall study design is shown in Figure 1. First, eligible patients were identified and allocated to the development and external validation cohorts according to the participating centers. Their preoperative multimodal data were collected and preprocessed. Second, modality-specific branches were constructed to extract complementary information from clinical information, static BUS and CDFI images, and dynamic DCE-US sequences. Third, heterogeneous representations were aligned and integrated through an attention-based multimodal fusion framework for MVI prediction. Finally, the contributions of individual modalities, modality combinations, fusion strategies, and key model components were systematically evaluated. Model discrimination, calibration, clinical net benefit, interpretability, and reader performance were also assessed.

## 2.2 Patients and cohort allocation

A total of 489 patients with pathologically confirmed HCC were enrolled from eight medical centers (Table S1) between July 2018 and April 2022. All patients underwent hepatectomy and had definitive postoperative MVI status. Histopathologic examination of the resected specimens served as the reference standard. MVI status was classified as M0 (no MVI), M1 (low-grade MVI), or M2 (high-grade MVI). For binary prediction, M1 and M2 were combined as MVI-positive, whereas M0 was considered MVI-negative.

Patients were allocated to the development and external validation cohorts. The development cohort comprised 421 patients from seven centers and was used for model development and internal validation through five-fold cross-validation. In each cross-validation iteration, the development cohort was divided into a training set for parameter optimization, a tuning set for hyperparameter selection and model selection, and an internal validation set for performance evaluation. The remaining 68 patients from eighth center constituted the external validation cohort and were not used for model training or hyperparameter optimization. The patient enrollment and cohort allocation process is shown in Figure S1.

The inclusion criteria were as follows: (1) age ≥18 years; (2) histopathologically confirmed HCC after hepatectomy, with definitive postoperative MVI status; (3) available clinical records and pathological reports; and (4) complete preoperative BUS, CDFI, and DCE-US examinations performed within one month before surgery. The exclusion criteria were: (1) corrupted imaging data or image quality inadequate for analysis; and (2) previous locoregional treatment, including transarterial chemoembolization or ablation.

## 2.3 Multimodal data acquisition and preprocessing

### 2.3.1 Clinical information

For each patient, 10 clinical variables and 22 manually assessed ultrasound features were collected. The clinical variables were extracted from medical records by clinicians and included age, sex, medical history, and laboratory measurements such as alpha-fetoprotein (AFP) and carcinoembryonic antigen (CEA) (Table 1). AFP was analyzed both as a continuous variable and as a categorical variable using established thresholds of 20 and 400 ng/mL [22].

The manually assessed ultrasound features comprised 14 conventional ultrasound features and eight DCE-US features, such as lesion size, enhancement characteristics, and perfusion timing parameters. Detailed definitions of the DCE-US features are provided in Supplementary Methods (Section 1), and all 22 features are summarized in Table S2. Lesion blood flow signals on CDFI were graded using the four-level Adler classification system [23]. The ultrasound features were independently evaluated using RadiAnt DICOM Viewer (version 4.6.9) by two abdominal ultrasound physicians (Q.Y. and L.D.), each with more than five years of experience in liver ultrasound and DCE-US interpretation. Both readers were blinded to the pathological MVI status and model predictions. Before formal assessment, they reviewed 10 cases that were not included in the study cohort to standardize the feature definitions and evaluation procedure. Disagreements were first resolved by consensus; if consensus could not be reached, a senior ultrasound physician (J.Y.) with 15 years of experience made the final determination.

### 2.3.2 Ultrasound

BUS, CDFI, and DCE-US examinations were performed using 12 ultrasound systems across the participating centers (Table S3). All examinations were conducted by ultrasound physicians who had received standardized training and had more than five years of experience in liver DCE-US. Image acquisition followed previously established examination protocols [22,24,25].

For each lesion, the BUS image showing the maximum lesion diameter and the corresponding CDFI image were selected. DCE-US was performed using a low mechanical index. After intravenous administration of 1.2–2.4 mL of SonoVue through an antecubital vein, the injection was followed by a 5-mL flush of 0.9% saline. Timing began at the initiation of contrast-agent injection. The target lesion and surrounding liver parenchyma were continuously observed for 2 min, and additional delayed-phase clips were subsequently acquired according to the established protocol.

The first 2 min of each DCE-US examination were used for temporal modeling because this interval captured the principal enhancement and washout differences between the lesion and surrounding liver parenchyma. The lesion region of interest was used as the imaging input for BUS, CDFI, and DCE-US. Detailed procedures for region-of-interest annotation, image de-identification, DCE-US video segmentation, and sequence-level region-of-interest generation are provided in Supplementary Materials (Section 2).

## 2.4 Development of the multimodal fusion model

The proposed MIFNN integrated clinical information with complementary representations derived from BUS, CDFI, and DCE-US. Because these modalities differ substantially in data structure and clinical meaning, the model first encoded each modality separately, then aligned the heterogeneous ultrasound representations before attention-based multimodal fusion. The overall architecture is illustrated in Figure 2.

MIFNN comprised a clinical information branch, two static-image branches, and a DCE-US temporal branch. In the clinical information branch, the clinical variables and manually assessed ultrasound features were converted into standardized text sequences and encoded using ClinicalBERT [26] to obtain a semantic representation of the clinical information. For static imaging, BUS and CDFI images were processed by ViT-B encoders pretrained using DINOv3 self-supervised learning [27,28]. The BUS branch was intended to characterize grayscale morphology, including lesion shape, margins, and internal echogenicity, whereas the CDFI branch captured intratumoral and peritumoral blood-flow patterns. In the DCE-US branch, the same ViT-B backbone was used to generate frame-level visual representations, which were subsequently processed by the proposed hemodynamic temporal change module (HTCM).

HTCM was designed to capture the temporal evolution of contrast-agent perfusion. Specifically, forward and backward differences between adjacent frame-level representations were used to characterize temporal changes during contrast arrival, enhancement, and washout. A temporal attention mechanism then emphasized informative perfusion phases and aggregated the resulting features into a sequence-level DCE-US representation. Through this process, the DCE-US branch jointly encoded spatial appearance and temporal perfusion dynamics.

BUS, CDFI, and DCE-US characterize different aspects of the tumor and therefore have heterogeneous feature distributions. Directly combining these representations may hinder the learning of consistent cross-modal associations. To reduce this heterogeneity before fusion, we introduced a representation consistency learning module (RCLM), using DCE-US as the alignment anchor because it most directly characterizes dynamic microvascular perfusion. RCLM applied bidirectional InfoNCE-based [29] contrastive objectives to the BUS–DCE-US and CDFI–DCE-US representation pairs. Representations from the same patient were treated as positive pairs, whereas those from different patients were treated as negative pairs. These objectives encouraged matched cross-modal representations to be more similar than unmatched representations while preserving information relevant to MVI prediction.

After representation alignment, the BUS, CDFI, and DCE-US representations were then augmented with learnable modality embeddings [30] and combined with a learnable classification token. These tokens were passed through a Transformer encoder to model interactions among the three ultrasound modalities and produce an integrated imaging representation. The imaging representation was subsequently combined with the clinical representation generated by ClinicalBERT, and a multilayer perceptron classification head produced the final probability of MVI. The model was optimized using a composite objective comprising binary cross-entropy loss for MVI classification and two cross-modal consistency losses for BUS–DCE-US and CDFI–DCE-US alignment. Details of the model architecture, loss function, and training procedures are provided in Supplementary Materials (Sections 3–5).

## 2.5 Modality comparisons and ablation experiments

Comparative experiments was conducted to evaluate the contribution of each input modality and the principal design choices of MIFNN. These experiments included single-modality modeling, multimodal combination analysis, fusion method comparison, and module ablation. Unless otherwise specified, all experiments used the same data partitions, preprocessing procedures, and performance evaluation criteria.

First, single-modality models were evaluated to determine the predictive value of clinical information, BUS, CDFI, and DCE-US. For clinical information modeling, ClinicalBERT was compared with logistic regression, XGBoost, support vector machine, random forest, multilayer perceptron, and graph convolutional network [31] models using the same input features. For BUS and CDFI, different image encoders and pretraining strategies were compared to determine the backbone for the multimodal framework. For DCE-US, random, uniform, and pixel-difference-based sampling were evaluated to determine the frame selection strategy. The selected frames were then modeled using representative three-dimensional video networks (including R3D [32], TimeSformer [33], and Swin3D [34]) and two-dimensional frame encoders with temporal aggregation methods (including ToShift [35], PTA [36], TCStyle [37], and the proposed HTCM).

Second, multimodal combinations were evaluated to examine the complementary value of different modalities. These combinations included BUS + CDFI, BUS + DCE-US, CDFI + DCE-US, BUS + CDFI + DCE-US, and the full model integrating all three ultrasound modalities with clinical information. Only the branches corresponding to the included modalities were retained, while all other settings and evaluation procedures remained unchanged.

Third, the attention-based fusion method was compared with element-wise addition [38], averaging, feature concatenation [39], bilinear fusion [40], and Attention Bottleneck [41]. All methods used the same modality-specific representations and differed only in their fusion mechanisms. Transformer encoders with different numbers of layers were also evaluated to determine the depth of the fusion module.

Finally, the individual and combined contributions of HTCM and RCLM were assessed through module ablation. Starting from a baseline model without either module, HTCM and RCLM were introduced separately and then jointly. All other network components, input modalities, and training settings were held constant.

## 2.6 Model interpretability

Post hoc interpretability analyses were performed for the imaging and clinical information branches. For the imaging branches, gradient-weighted class activation mapping (Grad-CAM) [42] was used to generate class-specific activation maps for BUS and CDFI images and individual DCE-US frames. The maps were superimposed on the corresponding images to visualize regions within the lesion ROI showing strong model activation. For DCE-US, maps were examined across the sampled frames to characterize changes in model activation during contrast arrival, enhancement, and washout. For the clinical information branch, the normalized attention weights of ClinicalBERT for each clinical feature were visualized to analyze their relative contribution to model prediction.

## 2.7 Reader study

A reader study was conducted to compare the performance of MIFNN with manual assessment. Two ultrasound physicians with different levels of experience in liver ultrasound independently evaluated all patients in the external validation cohort. For each patient, the readers reviewed the BUS and CDFI images, the DCE-US sequence, and the clinical variables and classified the case as MVI-positive or MVI-negative. Both readers were blinded to the histopathologic reference standard and model predictions.

## 2.8 Statistical analysis

Univariable analyses were performed to examine the associations of clinical variables and manually assessed ultrasound features with MVI status. Continuous variables were compared using the Mann–Whitney U test, and categorical variables were compared using the chi-square test or Fisher’s exact test, as appropriate.

Model discrimination was evaluated primarily using AUC. Accuracy (ACC), sensitivity (SEN), specificity (SPE), and F1 score were also calculated. For each cross-validation fold, the classification threshold was determined by maximizing the Youden index in the corresponding tuning set. This threshold was subsequently applied to the internal validation set and external validation cohort. Performance metrics were summarized as the mean and standard deviation across the five folds.

Calibration curves were used to assess agreement between predicted probabilities and observed outcomes. Decision curve analysis was performed to quantify clinical net benefit across a range of threshold probabilities relative to treat-all and treat-none strategies. Inter-reader agreement in the reader study was assessed using Cohen's κ coefficient. All statistical tests were two-sided, and $P < 0.05$ was considered statistically significant. Statistical analyses were performed using SciPy (version 1.13.1) and statsmodels (version 0.14.6) in Python.

## 3. Results

### 3.1 Associations of clinical information with MVI

Ten clinical variables and 22 manually assessed ultrasound features were individually evaluated for their associations with MVI status. Ten features were found to be significantly associated with MVI status (Table S4): history of chronic disease, overall enhancement pattern, late phase degree of enhancement, necrosis, AFP classification, maximum diameter of lesion, start time of washout, rise time, AFP level, and tumor infiltration boundary. These features were subsequently used as inputs to the clinical information models. Among them, maximum diameter of lesion, start time of washout, AFP level, and tumor infiltration boundary showed the strongest statistical evidence of association (all $P < 0.001$).

### 3.2 Performance of multimodal fusion

Table 2 summarizes the performance of different modality combinations in the external validation cohort. DCE-US was the best-performing single modality, with an AUC of $0.8545 \pm 0.0198$. Combining DCE-US with BUS or CDFI increased the AUC to $0.8670 \pm 0.0164$ and $0.8720 \pm 0.0263$, respectively, whereas combining BUS and CDFI without DCE-US yielded a substantially lower AUC of $0.6679 \pm 0.0492$.

Integration of all three ultrasound modalities further increased the AUC to 0.8860 ± 0.0121. The full multimodal model, which additionally incorporated clinical information, achieved the highest AUC (0.8953 ± 0.0180) and accuracy (81.18% ± 2.83%). Although its specificity (78.14% ± 6.28%) was lower than that of the three-ultrasound model (83.26% ± 8.61%), the full model achieved higher sensitivity for identifying MVI-positive cases.

Figure 3 shows the fold-specific ROC curves of the full multimodal model in the training set, internal validation set, and external validation cohort. AUCs ranged from 0.9563 to 0.9973 in the training sets and from 0.7155 to 0.8666 in the internal validation sets. In the external validation cohort, the model achieved consistently high discrimination across all five folds, with AUCs ranging from 0.8716 to 0.9153.

Figure 4 presents the calibration curves, decision curve analyses, and confusion matrices of the full multimodal model from the first fold. The calibration curves showed generally good agreement between predicted probabilities and observed outcomes, although some deviations were observed in parts of the probability range. Decision curve analysis showed greater net benefit for the model than for the treat-all and treat-none strategies across a wide range of threshold probabilities. In the external validation cohort, the model correctly identified 23 of 25 MVI-positive patients and 33 of 43 MVI-negative patients using the threshold determined in the corresponding tuning set.

## 3.3 Comparative performance of single-modality models

### 3.3.1 Clinical information models

The performance of the clinical information models is summarized in Table S5. ClinicalBERT achieved an AUC (0.6715 ± 0.0156), followed by logistic regression (0.6491 ± 0.0376) and support vector machine (0.6392 ± 0.0253). XGBoost and random forest yielded similar AUCs of 0.6180 ± 0.0471 and 0.6172 ± 0.0409, respectively, whereas multilayer perceptron and graph convolutional network showed lower discrimination.

### 3.3.2 BUS and CDFI image models

Table S6 summarizes the performance of the evaluated image encoders and pretraining strategies for BUS and CDFI. Overall, the CDFI models generally yielded higher AUCs than the BUS models. ViT-B pretrained with DINOv3 achieved the highest AUC for both modalities, reaching $0.6087 \pm 0.0417$ for BUS and $0.6435 \pm 0.0344$ for CDFI in the external validation cohort.

### 3.3.3 DCE-US temporal models

Table S7 compares the performance of three DCE-US frame sampling strategies. Pixel-difference-based sampling achieved an AUC of 0.8521, compared with 0.8326 for uniform sampling and 0.7963 for random sampling. These results suggest that preferentially sampling frames with greater inter-frame intensity changes may better preserve perfusion-related temporal information than random or uniform sampling.

Table 3 compares the performance of different DCE-US video modeling methods. The proposed HTCM yielded an AUC of $0.8545 \pm 0.0198$, whereas three-dimensional video networks (R3D [32], TimeSformer [33], and Swin3D [34]) and other two-dimensional frame-based temporal models (ToShift [35], PTA [36], and TCStyle [37]) showed lower AUCs. HTCM contained 87.51 million parameters, fewer than the other evaluated two-dimensional temporal models, while maintaining comparable computational complexity. Taken together, the frame-sampling and temporal-modeling results support the effectiveness of explicitly capturing dynamic perfusion changes in DCE-US for MVI prediction.

## 3.4 Fusion strategy comparison and ablation analysis

Table S8 compares the strategies used to integrate BUS, CDFI, and DCE-US

representations. Simple addition, averaging, concatenation, and bilinear fusion achieved AUCs ranging from 0.7801 to 0.8305. Attention-based methods generally yielded higher AUCs, with Attention Bottleneck achieving an AUC of 0.8655 ± 0.0244. Among the Transformer configurations, the two-layer encoder achieved the highest AUC (0.8860 ± 0.0121), compared with 0.8681 ± 0.0185 and 0.8759 ± 0.0174 for the one- and three-layer encoders, respectively.

The ablation results are presented in Table 4. The baseline model without HTCM or RCLM achieved an AUC of 0.8355 ± 0.0440. Adding HTCM or RCLM individually increased the AUC to 0.8629 ± 0.0352 and 0.8675 ± 0.0584, respectively. Incorporating both modules yielded the highest AUC (0.8953 ± 0.0180). These findings indicate that both temporal perfusion modeling and cross-modal representation alignment contributed to the observed performance gains, with their combination providing the best performance.

## 3.5 Visual interpretability of the model

Figure 5 presents the activation maps of the BUS, CDFI, and DCE-US branches, together with the ClinicalBERT attention distributions, for three representative cases. Across the imaging modalities, regions of high activation were predominantly located within the lesions or along their margins. The DCE-US maps further showed that model activation varied across enhancement phases, indicating that the temporal branch attended to changing perfusion patterns throughout the sequence.

In the correctly classified MVI-positive case (Figure 5A; predicted probability, 0.828), high activation was observed in the tumor parenchyma and marginal regions on BUS, in local blood-flow regions on CDFI, and across multiple phases of the DCE-US sequence. Among the clinical information inputs, start time of washout, AFP level, and tumor infiltration boundary received relatively high attention weights. In the correctly classified MVI-negative case (Figure 5B; predicted probability, 0.075), imaging activation was generally less concentrated across the three ultrasound modalities, and the distribution of clinical attention weights differed from that in the MVI-positive case.

The false-positive case (Figure 5C; predicted probability, 0.557) showed prominent activation in lesion-related regions on BUS and CDFI and across several DCE-US phases. Several clinical information inputs, including maximum diameter of lesion, rise time, AFP level, start time of washout, and tumor infiltration boundary, also received relatively high attention weights. These patterns resembled those observed in the correctly classified MVI-positive case and may have contributed to the false-positive prediction. Overall, the visualizations indicate that the model focused primarily on lesion-related morphology, blood flow, temporal perfusion patterns, and specific clinical features when generating predictions.

## 3.6 Reader performance

The results of the reader study in the external validation cohort are summarized in Table S9. Reader 1 achieved an accuracy of 54.41%, sensitivity of 64.00%, specificity of 48.84%, and F1 score of 50.79%. The corresponding values for Reader 2 were 44.62%, 64.00%, 32.50%, and 47.06%, respectively. Inter-reader agreement was moderate (Cohen's $\kappa = 0.4202$). Both readers showed lower performance than the full multimodal model in the same cohort. These findings suggest that manual integration of multimodal ultrasound and clinical information for preoperative MVI assessment is challenging.

# 4. Discussion

In this multicenter study, we developed a multimodal fusion framework for preoperative prediction of MVI in patients with HCC. DCE-US provided substantially greater discriminative value than BUS, CDFI, or clinical information alone, highlighting the importance of dynamic perfusion characteristics for MVI assessment. Integrating the complementary ultrasound modalities and clinical information further improved predictive performance, with the full multimodal model achieving an AUC of $0.8953 \pm 0.0180$ in the external validation cohort. Moreover, HTCM and RCLM each improved model performance, with the greatest improvement observed when both modules were incorporated. These findings support the complementary value of temporal perfusion modeling and cross-modal representation alignment.

The strong performance of DCE-US is biologically plausible because MVI is associated with alterations in tumor microvasculature and the surrounding vascular microenvironment. Whereas BUS primarily depicts tumor morphology and CDFI visualizes macroscopic blood flow, DCE-US captures the temporal evolution of contrast arrival, enhancement, peak intensity, and washout, providing a more direct representation of tumor perfusion abnormalities. Pixel-difference-based sampling achieved a higher AUC than random or uniform sampling, suggesting that frames showing greater signal changes retain more informative perfusion dynamics. By explicitly modeling bidirectional changes between adjacent frames, HTCM achieved a higher AUC than the evaluated general-purpose video and temporal models. These findings support the value of physiologically informed temporal modeling for DCE-US analysis.

MVI-related phenotypes extend beyond dynamic perfusion. BUS and CDFI provide complementary information on tumor morphology, margins, internal structure, and macroscopic vascularity, while clinical information provides additional context patient and tumor characteristics. ClinicalBERT achieved a higher AUC than the other clinical information models, suggesting that contextual representation learning may capture relationships among these features more effectively than conventional tabular modeling. RCLM was designed to reduce representation heterogeneity across the ultrasound modalities before Transformer-based fusion. Accordingly, the full multimodal model achieved higher AUCs than the individual modalities and their partial combinations. Previous ultrasound-based studies have primarily used DCE-US alone, limited modality combinations, manually engineered features, or conventional statistical models [15–21]. The present framework extends this work by integrating multimodal ultrasound and clinical information with explicit temporal modeling and representation alignment and by evaluating performance in an independent cohort from an eighth center.

The visualization results provided additional context for the model predictions. Activation maps predominantly highlighted the lesions, tumor margins, vascular regions, and areas showing dynamic enhancement changes, while the clinical information branch assigned relatively high weights to clinical features such as AFP level, start time of washout, and tumor infiltration boundary. These patterns are broadly consistent with characteristics associated with aggressive HCC phenotypes. In the reader study, both ultrasound physicians showed lower classification performance than the full multimodal model, illustrating the difficulty of manually integrating heterogeneous imaging and clinical information for preoperative MVI assessment.

This study has several limitations. First, despite the multicenter development and validation design, larger prospective studies involving additional institutions are warranted to further establish the model's generalizability. Second, the framework requires complete multimodal inputs from BUS, CDFI, DCE-US, and clinical information. Future work should therefore address missing modalities and improve robustness under incomplete-input conditions. Third, the present study focused on binary prediction of pathological MVI status. Longitudinal studies are needed to determine whether model-derived risk estimates are associated with recurrence, survival, or treatment benefit.

## 5. Conclusions

The proposed framework integrates tumor morphology, vascularity, dynamic perfusion, and clinical information for preoperative MVI prediction in HCC. Hemodynamics-aware temporal modeling effectively captures perfusion changes in DCE-US, while cross-modal representation alignment and multimodal fusion further exploit the complementary value of BUS, CDFI, and clinical information. These findings support the potential of the proposed framework as an auxiliary tool for preoperative MVI risk stratification and individualized clinical decision-making in patients with HCC.

## List of abbreviations

MVI: Microvascular invasion

HCC: Hepatocellular carcinoma

BUS: B-mode ultrasound

CDFI: Color Doppler flow imaging

DCE-US: Dynamic contrast-enhanced ultrasound

AUC: Area under the receiver operating characteristic curve

MIFNN: multimodal information fusion neural network

AFP: Alpha-fetoprotein

CEA: Carcinoembryonic antigen

HTCM: hemodynamic temporal change module

RCLM: representation consistency learning module

Grad-CAM: gradient-weighted class activation mapping

ACC: Accuracy

SEN: sensitivity

SPE: specificity

## Declarations

### Ethics approval and consent to participate

This study was approved by the ethics committees of all participating centers and registered on ClinicalTrials.gov (NCT03871140). All patients provided written informed consent. Personal health information was de-identified.

### Consent for publication

Not applicable.

### Data and software availability

The datasets generated and analyzed during this study are not publicly available due to patient privacy and institutional restrictions but are available from the corresponding author on reasonable request. The code used for data analysis and model development is publicly available on GitHub at https://github.com/kyy147/MIFNN.

### Competing interests

The authors declare that they have no competing interests.

### Funding

This work was supported by Shenzhen Medical Research Fund (No. D2602009), Guangdong Basic and Applied Basic Research Foundation (Nos. 2025A1515011821 and 2026A1515012488), Shenzhen Science and Technology Program (No. JCYJ20240813143302004), National Natural Science Foundation of China (Nos.

12326619 and 62171290), Science and Technology Planning Project of Guangdong Province (No. 2023A0505020002), and Frontier Technology Development Program of Jiangsu Province (No. BF2024078).

**Author's contributions**

JC: conceptualization, methodology, formal analysis, writing – original draft and writing – review & editing. YK and QH: methodology, formal analysis, validation, and writing – original draft. XT, LD, YH, WX, RH and DN: data curation, investigation, visualization, and resources. QY, JY and PL: conceptualization, project administration, resources, supervision and writing – review & editing. All authors read and approved the final manuscript.

**Acknowledgements**

Not applicable.

# Figures

**A Model development using five-fold cross-validation in the development cohort (7 centers; n = 421)**

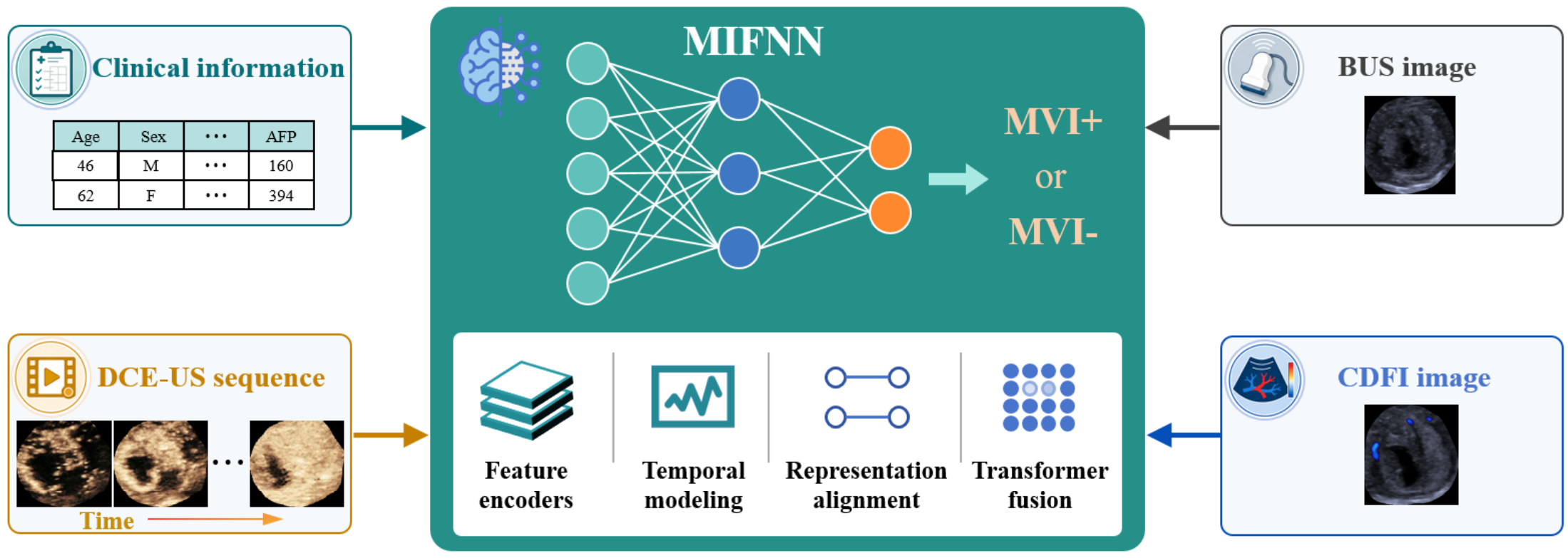

**B Model evaluation in the independent external validation cohort (1 center; n=68)**

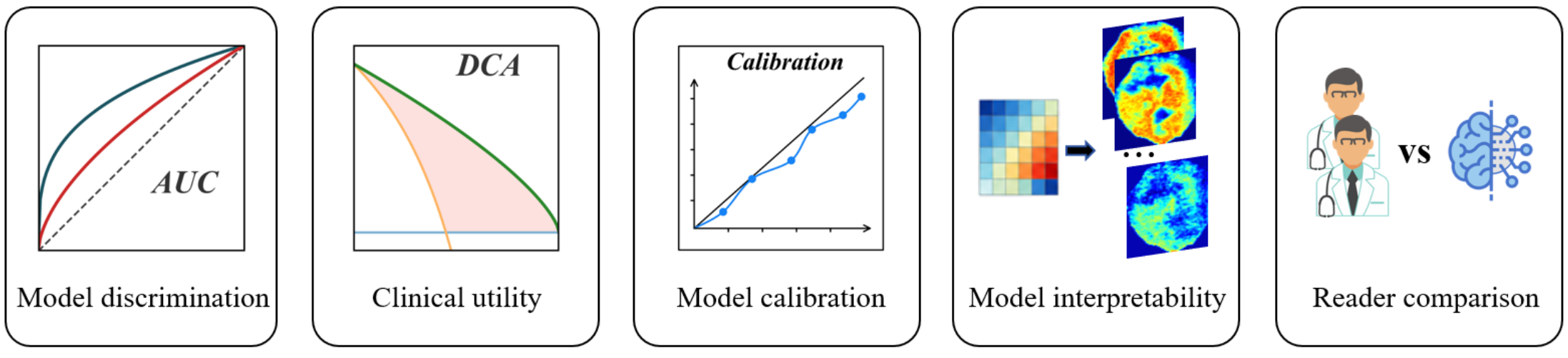

**Figure 1.** Overall workflow of model development and evaluation. (A) Clinical information, BUS images, CDFI images, and DCE-US sequences were processed using modality-specific branches. The extracted representations underwent temporal modeling, representation alignment, and Transformer-based fusion to generate MVI positive or negative predictions. (B) Independent evaluation in the external validation cohort from an eighth center, including assessment of model discrimination, clinical utility, calibration, interpretability, and reader performance. BUS, B-mode ultrasound; CDFI, color Doppler flow imaging; DCE-US, dynamic contrast-enhanced ultrasound; MIFNN, multimodal information fusion neural network; MVI, microvascular invasion.

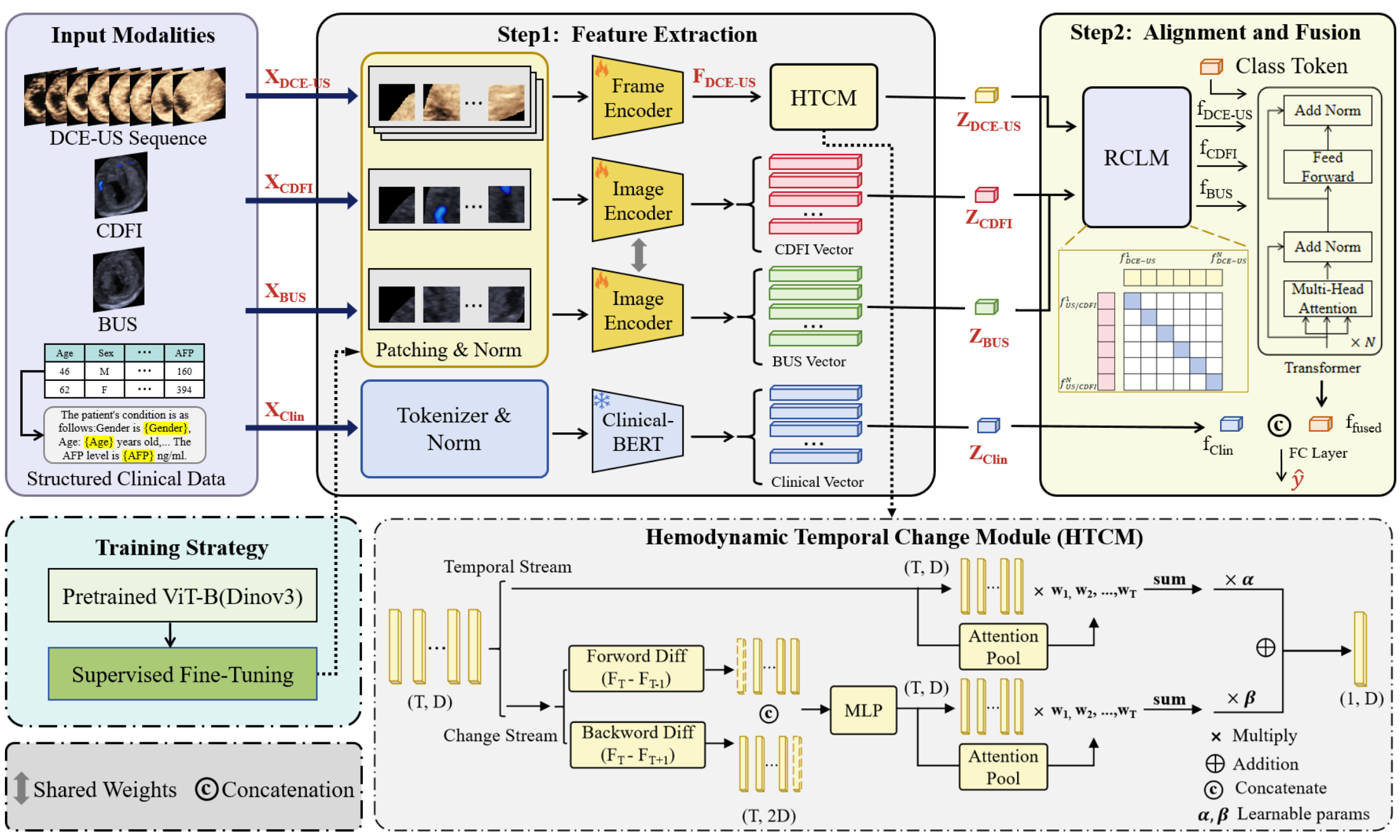

**Figure 2.** Architecture of the proposed MIFNN for preoperative MVI prediction. Multimodal ultrasound and clinical data were processed using modality-specific encoders. The HTCM modeled bidirectional inter-frame changes in the DCE-US sequence, and the RCLM aligned the representations of the three ultrasound modalities. The aligned ultrasound representations were subsequently integrated using a Transformer-based fusion module. The resulting multimodal ultrasound representation was concatenated with the clinical representation and passed through a fully connected layer to generate the MVI prediction. BUS, B-mode ultrasound; CDFI, color Doppler flow imaging; DCE-US, dynamic contrast-enhanced ultrasound; HTCM, hemodynamic temporal change module; MIFNN, multimodal information fusion neural network; MVI, microvascular invasion; RCLM, representation consistency learning module.

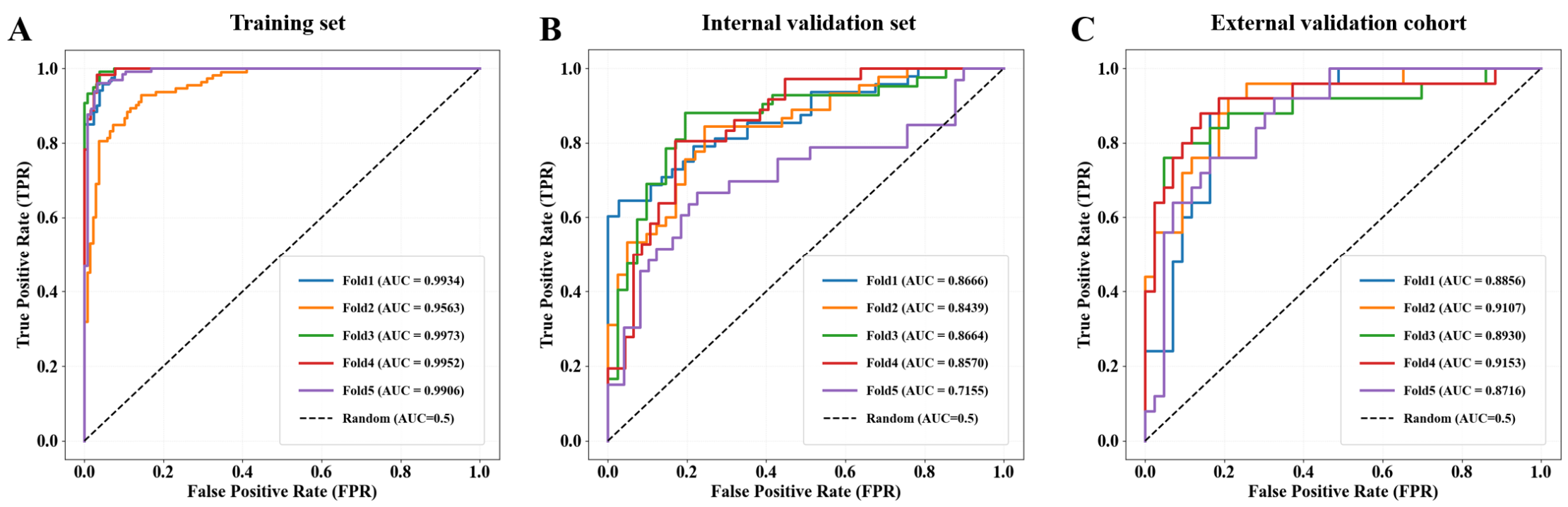

**Figure 3.** ROC curves of the full multimodal model across five cross-validation folds. ROC curves are shown for the training sets (A), internal validation sets (B), and independent external validation cohort (C). AUC, area under the receiver operating characteristic curve; ROC, receiver operating characteristic.

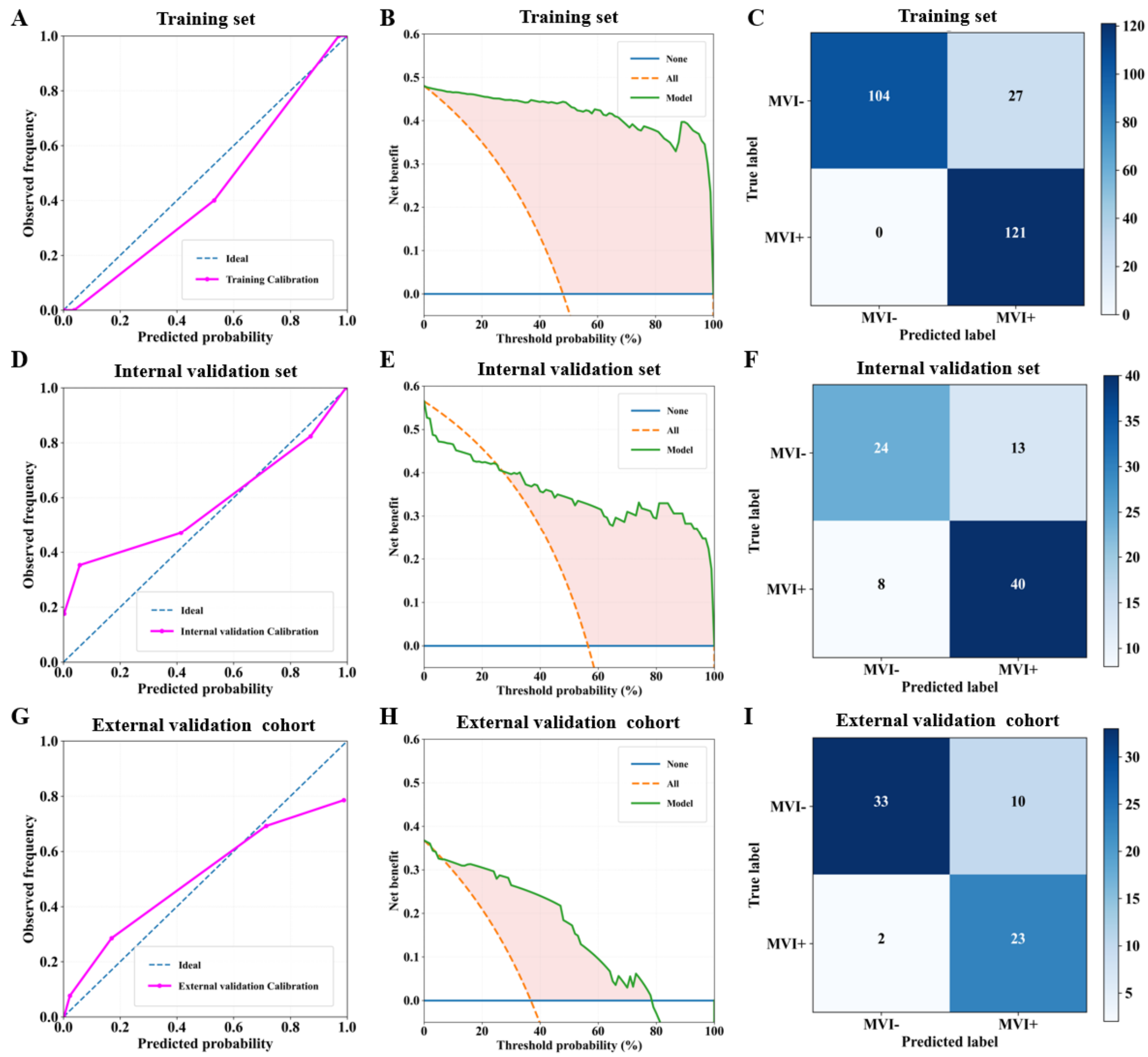

**Figure 4.** Calibration curves, decision curve analyses, and confusion matrices of the full multimodal model in Fold 1 are shown for the training set (A-C), internal validation set (D-F), and external validation cohort (G-I). MVI, microvascular invasion.

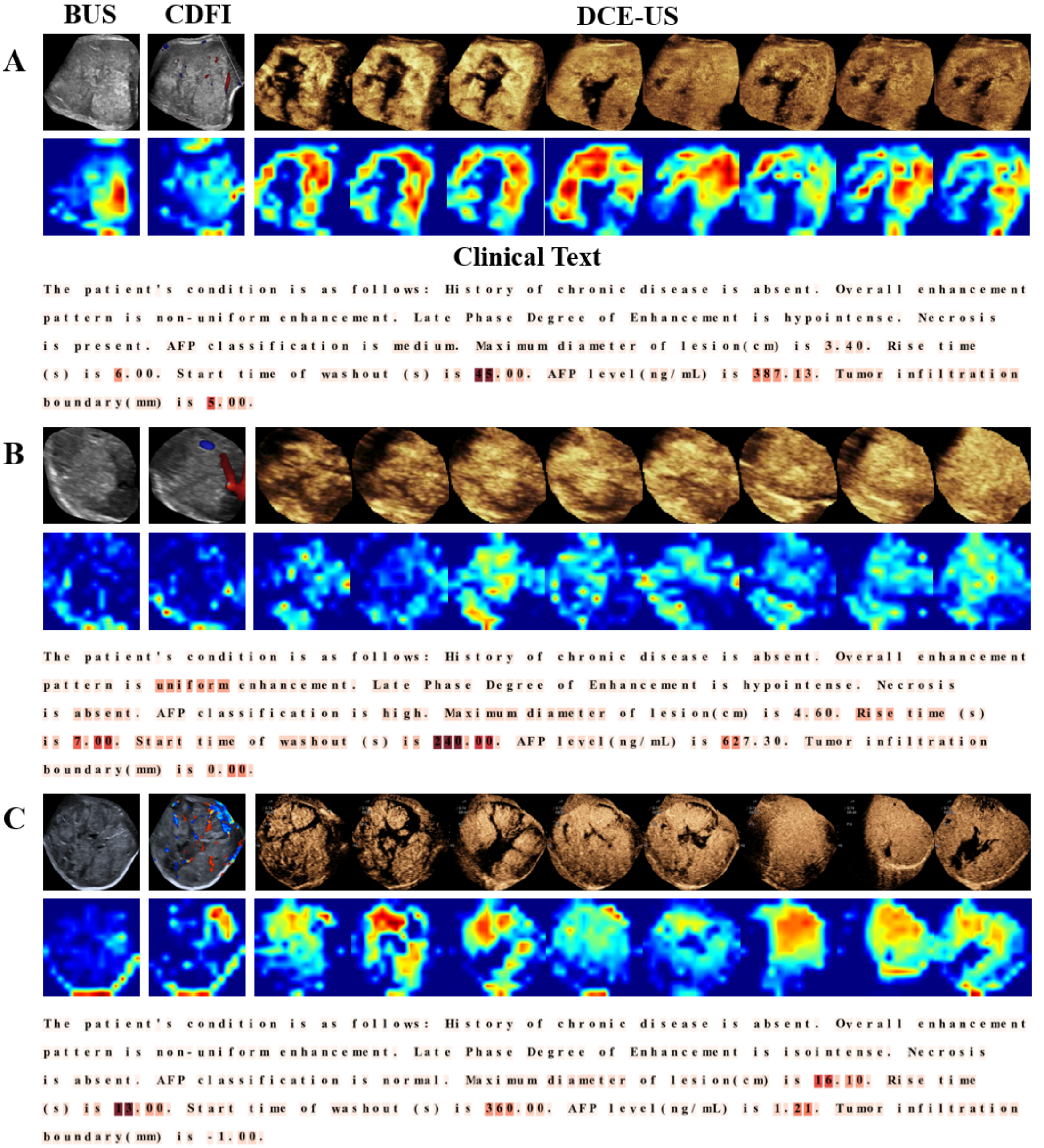

**Figure 5.** Model interpretability visualizations for three representative cases. Ultrasound images and corresponding saliency maps are shown together with the attention weights assigned to the clinical text by the full multimodal model. (A) Correctly classified MVI-positive case (predicted probability, 0.828). (B) Correctly classified MVI-negative case (predicted probability, 0.075). (C) False-positive case (predicted probability, 0.557). BUS, B-mode ultrasound; CDFI, color Doppler flow imaging; DCE-US, dynamic contrast-enhanced ultrasound; MVI, microvascular invasion.

# Tables

**Table 1.** Patient characteristics in the development and external validation cohorts.

| **Characteristics** | **All patients (n = 489)** | **Development cohort (n = 421)** | **External validation cohort (n = 68)** |
|---|---|---|---|
| Sex | | | |
| Male | 408 (83.4%) | 352 (83.6%) | 56 (82.4%) |
| Female | 81 (16.6%) | 69 (16.4%) | 12 (17.6%) |
| Age (years) | 55.4 (26.0-86.0) | 55.4 (26.0-83.0) | 55.5 (30.0-86.0) |
| AFP level (ng/mL) | 3077.0 ($0.4\text{-}2.5\times10^5$) | 3407.6 ($0.8\text{-}2.5\times10^5$) | 1069.0 ($0.4\text{-}4.9\times10^4$) |
| CEA ( μ g/L) | 2.7 (0.0-25.0) | 2.7 (0.0-25.0) | 1.7 (0.4-3.0) |
| Underlying liver disease | | | |
| None | 59 (12.1%) | 48 (11.4%) | 11 (16.2%) |
| Hepatitis | 419 (85.7%) | 362 (86.0%) | 57 (83.8%) |
| Alcohol-associated | 11 (2.2%) | 11 (2.6%) | 0 |
| Clinical symptoms | | | |
| No | 389 (79.6%) | 334 (79.3%) | 55 (80.9%) |
| Yes | 100 (20.4%) | 87 (20.7%) | 13 (19.1%) |
| History of chronic diseases | | | |
| No | 372 (76.1%) | 315 (74.8%) | 57 (83.8%) |
| Yes | 117 (23.9%) | 106 (25.2%) | 11 (16.2%) |
| History of primary tumor | | | |
| Yes | 453 (92.6%) | 391 (92.9%) | 62 (91.2%) |
| No | 36 (7.4%) | 30 (7.1%) | 6 (8.8%) |
| History of malignant diseases | | | |
| No | 451 (92.2%) | 384 (91.2%) | 67 (98.5%) |
| Yes | 38 (7.8%) | 37 (8.8%) | 1 (1.5%) |
| AFP classification | | | |
| < 20ng/mL | 204 (42.4%) | 182 (44.1%) | 22 (32.4%) |
| 20~400ng/mL | 130 (27.0%) | 100 (24.2%) | 30 (44.1%) |
| > 400ng/mL | 147 (30.6%) | 131 (31.7%) | 16 (23.5%) |

Note. Discrete variables are represented as: number of cases (%) and continuous variables are represented as: median (minimum value - maximum value). AFP = alpha-fetoprotein; CEA = carcinoembryonic antigen.

**Table 2.** External validation performance of models based on individual and combined modalities.

| Modality | AUC | ACC (%) | SEN (%) | SPE (%) | F1 (%) |
|---|---|---|---|---|---|
| BUS | 0.6087±0.0417 | 58.24±5.75 | 59.20±21.05 | 57.67±19.83 | 49.93±7.00 |
| CDFI | 0.6435±0.0344 | 57.94±7.25 | 65.60±19.51 | 53.49±18.09 | 52.67±7.80 |
| DCE-US | 0.8545±0.0198 | 74.71±3.19 | 81.60±7.27 | 70.70±8.79 | 70.38±1.55 |
| Clinical information | 0.6715±0.0156 | 59.41±6.94 | 60.00±17.45 | 59.02±21.50 | 53.50±5.14 |
| BUS + CDFI | 0.6679±0.0492 | 55.88±6.41 | 73.60±22.56 | 45.58±22.58 | 54.26±5.24 |
| CDFI + DCE-US | 0.8720±0.0263 | 77.35±7.32 | 81.60±10.81 | 74.88±14.84 | 72.89±6.05 |
| BUS + DCE-US | 0.8670±0.0164 | 78.53±2.67 | 76.80±10.35 | 79.53±7.78 | 72.33±3.47 |
| BUS + CDFI + DCE-US | 0.8860±0.0121 | 79.41±1.80 | 72.80±10.73 | 83.26±8.61 | 72.05±2.26 |
| All modalities | **0.8953±0.0180** | 81.18±2.83 | 86.40±6.69 | 78.14±6.28 | 77.16±2.74 |

**Table 3.** Performance comparison of three-dimensional video models and two-dimensional frame-based temporal models for DCE-US-based MVI prediction in the external validation cohort.

| Data type | Model | AUC | ACC (%) | SEN (%) | SPE (%) | F1 (%) | Params | FLOPs |
|---|---|---|---|---|---|---|---|---|
| 3D | R3D | 0.6048±<br>0.0694 | 55.59±<br>13.09 | 60.00±<br>15.49 | 53.02±<br>27.63 | 50.14±<br>6.41 | 473.54K | 6.37G |
| | Swin3D | 0.8076±<br>0.0127 | 70.29±<br>4.92 | 69.60±<br>13.45 | 70.70±<br>15.04 | 63.12±<br>2.61 | 19.25M | 23.94G |
| | TimeSformer | 0.8413±<br>0.0118 | 71.18±<br>3.22 | 73.60±<br>19.92 | 69.77±<br>16.44 | 59.89±<br>19.37 | 121.50M | 379.81G |
| 2D | ViT-B<br>+ToShift | 0.8182±<br>0.0251 | 71.76±<br>2.83 | 72.00±<br>7.48 | 71.63±<br>8.29 | 65.20±<br>1.62 | 89.73M | 275.26G |
| | ViT-B<br>+PTA | 0.8287±<br>0.0244 | 72.94±<br>4.24 | 72.80±<br>10.35 | 73.02±<br>11.70 | 66.45±<br>2.59 | 91.22M | 275.28G |
| | ViT-B<br>+TCStyle | 0.8329±<br>0.0242 | 72.35±<br>4.21 | 72.80±<br>7.16 | 72.09±<br>7.54 | 65.96±<br>4.35 | 95.49M | 275.27G |
| | ViT-B<br>+HTCM | **0.8545±**<br>**0.0198** | 74.71±<br>3.19 | 81.60±<br>7.27 | 70.70±<br>8.79 | 70.38±<br>1.55 | 87.51M | 275.28G |

**Table 4.** Ablation analysis of HTCM and RCLM in the external validation cohort.

| HTCM | RCLM | AUC | ACC (%) | SEN (%) | SPE (%) | F1 (%) |
|---|---|---|---|---|---|---|
| | | 0.8355±<br>0.0440 | 74.71±<br>5.64 | 74.40±<br>14.31 | 74.88±<br>9.78 | 68.09±<br>7.70 |
| √ | | 0.8629±<br>0.0352 | 79.41±<br>3.60 | 74.40±<br>13.15 | 82.33±<br>6.06 | 72.28±<br>6.23 |
| | √ | 0.8675±<br>0.0584 | 77.65±<br>6.19 | 76.80±<br>12.46 | 78.14±<br>7.82 | 71.47±<br>8.64 |
| √ | √ | **0.8953±**<br>**0.0180** | 81.18±<br>2.83 | 86.40±<br>6.69 | 78.14±<br>6.28 | 77.16±<br>2.74 |

# Supplementary Materials

## Preoperative Prediction of Microvascular Invasion in Hepatocellular Carcinoma by Integrating Multimodal Ultrasound and Clinical Data: A Multicenter Study

## Contents

# Supplementary Methods

## 1. Manually assessed DCE-US imaging features

The following eight DCE-US features were assessed: (1) overall enhancement pattern, classified as heterogeneous enhancement, homogeneous enhancement, and no enhancement in all three phases; (2) portal venous phase degree of enhancement, classified as isoenhancement, hypoenhancement, and hyperenhancement, with the surrounding liver parenchyma as the reference; (3) late phase degree of enhancement, classified as isoenhancement and hypoenhancement, with the surrounding liver parenchyma as the reference; (4) start time of arterial phase enhancement (s); (5) time to peak (s), defined as the interval from contrast-agent injection to peak lesion enhancement; (6) rise time (s), defined as the interval from the start of arterial-phase lesion enhancement to peak enhancement; (7) start time of washout (s); and (8) tumor infiltration boundary, defined as the lesion size assessed by DCE-US minus the lesion size assessed by BUS, measured in millimeters.

## 2. Image annotation and segmentation

All multimodal data from the same patient were linked using a unique anonymized identifier. The original DICOM files were converted into PNG images for BUS and CDFI and AVI videos for DCE-US. For BUS and CDFI images, a largest connected-component algorithm was used to identify the fan-shaped ultrasound field of view, after which patient information and other irrelevant text outside the imaging region were removed (Figure S2). The tumor region of interest (ROI) was independently delineated by two experienced ultrasound physicians. Based on the resulting annotations, each ROI was expanded outward by 20 pixels to incorporate the peritumoral region (Figure S3). A binary lesion mask was subsequently generated from each contour, and the image was cropped to the minimum bounding rectangle enclosing the expanded mask.

DCE-US videos were displayed in a dual-panel format comprising spatially corresponding grayscale and contrast-enhanced ultrasound panels. The two panels were separated using a gradient-based boundary detection method and cropped to identical dimensions for subsequent analysis. Each segmentation result was manually reviewed to ensure spatial correspondence between the two panels. When inconsistencies in panel dimensions or alignment were detected, the cropping boundaries were manually corrected before further processing.

Because tumor boundaries may become indistinct in the contrast-enhanced panel after contrast-agent arrival, direct delineation on this panel may introduce localization errors. Therefore, two experienced ultrasound physicians selected representative key frames and delineated the tumor contours on the corresponding grayscale panel. The resulting contours were transferred to the spatially aligned contrast-enhanced panel to localize the tumor consistently across the two panels.

For each DCE-US sequence, the tumor contours delineated on the key frames were propagated to the corresponding locations in the remaining frames and manually adjusted when necessary to account for motion or changes in lesion appearance. To incorporate peritumoral tissue, the width and height of the rectangular binary mask enclosing each propagated tumor contour were each increased by 10%, with the mask center kept unchanged (Figure S4). All annotations were reviewed by a third experienced ultrasound physician. Any discrepancies were resolved through discussion until consensus was reached. This delineation–propagation–adjustment procedure was used to obtain ROIs throughout the complete DCE-US sequence while reducing the burden of frame-by-frame manual annotation.

## 3. Configuration of the modality-specific encoders

For clinical information modeling, we adopted ClinicalBERT, as shown in Figure

S5. ClinicalBERT follows the standard BERT-Base architecture, including 12 bidirectional Transformer encoder layers, a hidden dimension of 768, and 12 attention heads, with approximately 110 million parameters. Multi-head self-attention is used to model contextual relationships within the input sequence. ClinicalBERT is initially pretrained using the standard BERT objectives, including masked language modeling and next-sentence prediction, and is subsequently pretrained on clinical corpora. This domain-specific pretraining enables the model to learn semantic representations that are better adapted to clinical information.

For two-dimensional images, we adopted a ViT-Base-patch16 encoder initialized with weights obtained through DINOv3 self-supervised pretraining, followed by supervised fine-tuning on our training data, as shown in Figure S6. The core self-supervised training strategy of DINOv3 follows and further strengthens the "label-free distillation" paradigm of DINO/DINOv2. During training, a student network and a teacher network with the same ViT backbone are maintained simultaneously. The parameters of the teacher network are not updated through backpropagation but are obtained by the exponential moving average of the student network parameters, thereby providing a more stable learning target for the student and effectively reducing the risk of representation collapse during self-supervised training. DINOv3 employs a multi-crop augmentation strategy that generates views of different scales and viewpoints from the same input image, including a small number of global views and multiple local views. The student network receives both global and local views to learn cross-scale consistent representations, whereas the teacher network processes only global views. Under the constraint of "local–global consistency", the model learns feature representations with improved semantic discriminability and scale robustness.

For DCE-US sequences, the modality-specific encoder combined a ViT frame encoder with the HTCM, as shown in Figure 2. The ViT encoder first extracts high-level semantic features from each sampled frame, transforming the original video into a sequence of frame-level representations. The HTCM then explicitly models hemodynamic changes between adjacent frames and applies temporal attention to weight and aggregate the most informative perfusion phases. This process produces a sequence-level DCE-US representation that captures discriminative hemodynamic information for MVI prediction.

## 4. Loss functions

Binary cross-entropy loss was used for MVI classification:

$$L_{cls} = -\frac{1}{N}\sum_{i=1}^{N}[y_i \, log(\hat{y}_i) + (1 - y_i) \, log(1 - \hat{y}_i)],$$

where $N$ denotes the batch size, $y_i$ is the ground-truth label, and $\hat{y}_i$ is the predicted probability of MVI for the i-th patient.

Inspired by the bidirectional symmetric contrastive learning strategy used in CLIP [1], we designed the RCLM to align representations across ultrasound modalities. An InfoNCE-based objective was applied in both alignment directions, with representations from the same patient treated as positive pairs and those from different patients treated as negative pairs. This objective encouraged matched cross-modal representations to be closer while maintaining discrimination between unmatched representations in a shared feature space. Because DCE-US achieved the best single-modality performance and provided the richest dynamic perfusion information, it was used as the reference modality. BUS and CDFI representations were therefore aligned separately with the DCE-US representation.

Let $z_i^B$, $z_i^C$, and $z_i^D$ denote the BUS, CDFI, and DCE-US representations, respectively, for the i-th patient. With BUS as the anchor and DCE-US as the contrastive modality, the alignment loss was defined as

$$L_{B\to D} = -\frac{1}{N}\sum_{i=1}^{N} log \frac{exp(sim(z_i^B, z_i^D)/\tau)}{\sum_{j=1}^{N} exp\left(sim(z_i^B, z_j^D)/\tau\right)},$$

where $sim(\cdot,\cdot)$ denotes cosine similarity and $\tau$ is the temperature parameter. The reverse-direction loss was defined as

$$L_{D\to B} = -\frac{1}{N}\sum_{i=1}^{N} log \frac{exp(sim(z_i^D, z_i^B)/\tau)}{\sum_{j=1}^{N} exp\left(sim(z_i^D, z_j^B)/\tau\right)}.$$

The bidirectional BUS–DCE-US alignment loss was calculated as

$$L_{BD} = \frac{1}{2}(L_{B\to D} + L_{D\to B}).$$

Similarly, the bidirectional CDFI–DCE-US alignment loss was defined as

$$L_{CD} = \frac{1}{2}(L_{C\to D} + L_{D\to C}).$$

The total training loss was

$$\mathrm{L} = L_{cls} + \lambda_1 L_{BD} + \lambda_2 L_{CD},$$

where $\lambda_1$ and $\lambda_2$ control the contributions of the two alignment objectives. Both coefficients were set to 0.1 based on performance on the tuning set.

## 5. Model training

Unless otherwise specified, all models were optimized using AdamW [2] with a weight decay of $1 \times 10^{-4}$. The learning rate was reduced by 10% every 10 epochs. Early stopping was triggered if the tuning loss did not decrease for five consecutive epochs, and the model with the highest AUC on the tuning set was retained. Within each cross-validation fold, the classification threshold was determined by maximizing the Youden index on the tuning set and was subsequently applied to the corresponding internal validation set and the external validation cohort. The batch size was set to 1 during inference.

The ClinicalBERT-based clinical information model was trained for up to 60 epochs with a batch size of 32 and an initial learning rate of $2 \times 10^{-4}$.

For the BUS and CDFI models, images were resized to 224 × 224 pixels. Data augmentation included random horizontal and vertical flipping, rotation, color jittering, Gaussian noise, and cropping. The images were then normalized to the range [−1, 1]. Each model was trained for up to 60 epochs with a batch size of 16. The learning rates were set to $1 \times 10^{-6}$ for the feature-extraction backbone and $1 \times 10^{-5}$ for the classification head.

For the DCE-US model, the sampled video frames were resized to 224 × 224 pixels. Data augmentation included random horizontal and vertical flipping, rotation, color jittering, Gaussian noise, cropping, and temporal reversal of the frame sequence. The model was trained for up to 100 epochs with a batch size of 16. The learning rates were set to $2 \times 10^{-6}$ for the frame encoder and $2 \times 10^{-5}$ for the HTCM and classification head.

The full multimodal model was optimized using AdamW with a weight decay of $1 \times 10^{-3}$. The ClinicalBERT encoder was kept frozen during training. The DCE-US branch was initialized using the weights of the pretrained DCE-US model, whereas the BUS and CDFI branches were initialized using the original DINOv3-pretrained weights to allow further adaptation to the cross-modal alignment objectives. The three ultrasound branches were fine-tuned with a learning rate of $2 \times 10^{-6}$, whereas the fusion Transformer and MLP classification head used a learning rate of $1 \times 10^{-5}$. The model was trained for up to 50 epochs with a batch size of 16. The temperature parameter $\tau$ was set to 0.07, and the alignment loss weights $\lambda_1$ and $\lambda_2$ were both set to 0.1.

## Supplementary Figures

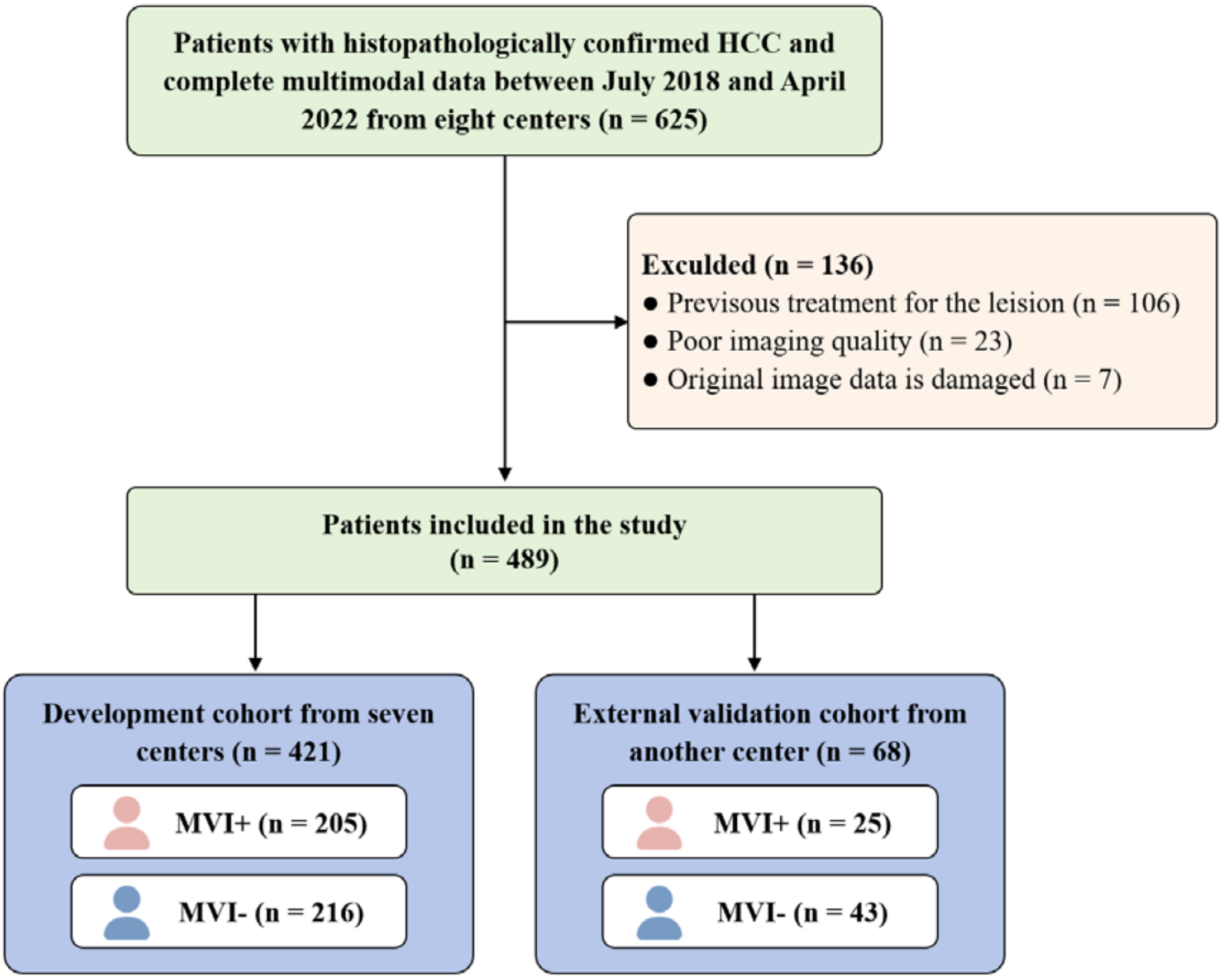

Figure S1. Patient enrollment and cohort allocation.

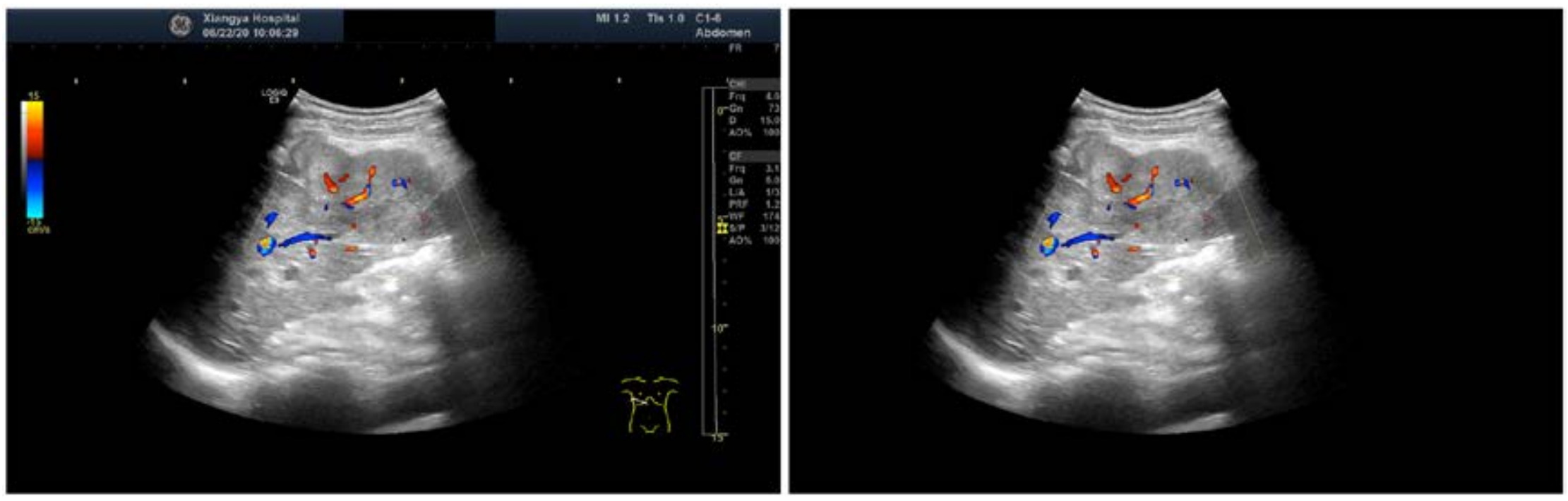

Figure S2. Extraction of the ultrasound field of view from a CDFI image. The original CDFI image is shown on the left, and the extracted ultrasound field of view after removal of irrelevant surrounding regions is shown on the right.

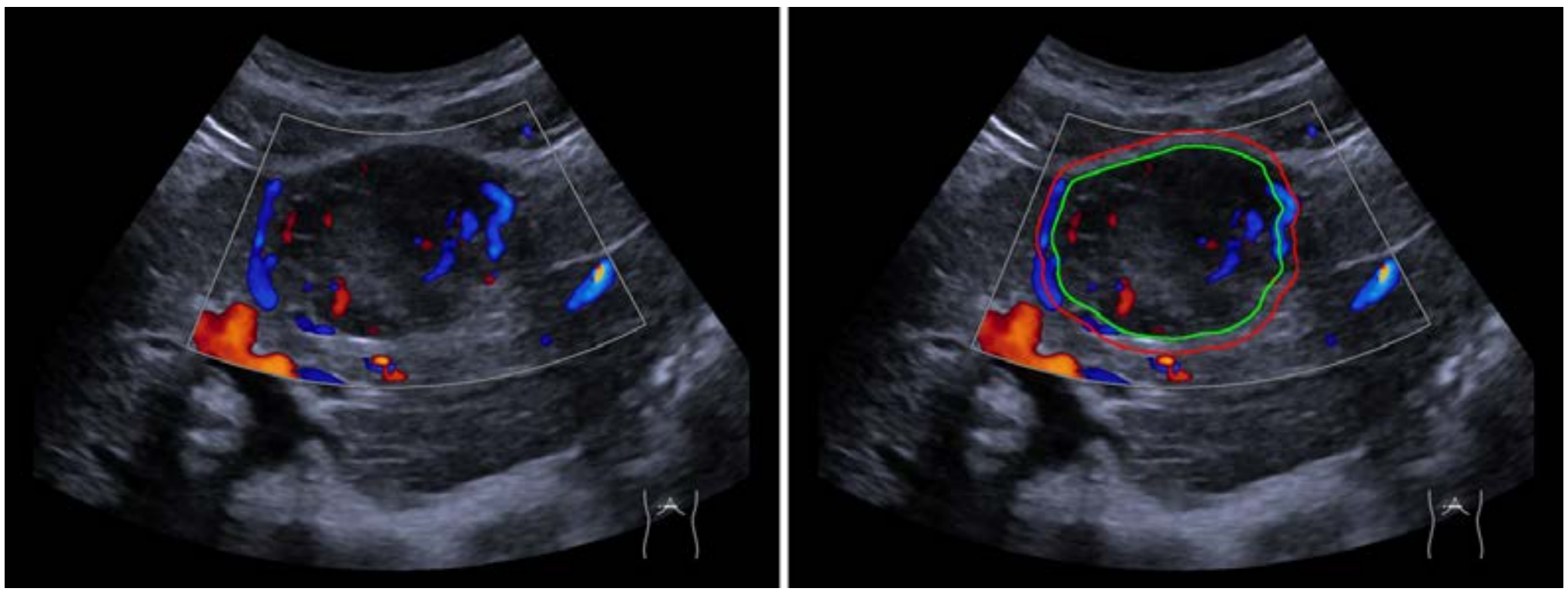

Figure S3. Expansion of the tumor region of interest in a CDFI image. The green contour indicates the original tumor annotation, and the red contour indicates the region of interest expanded outward by 20 pixels to include peritumoral tissue.

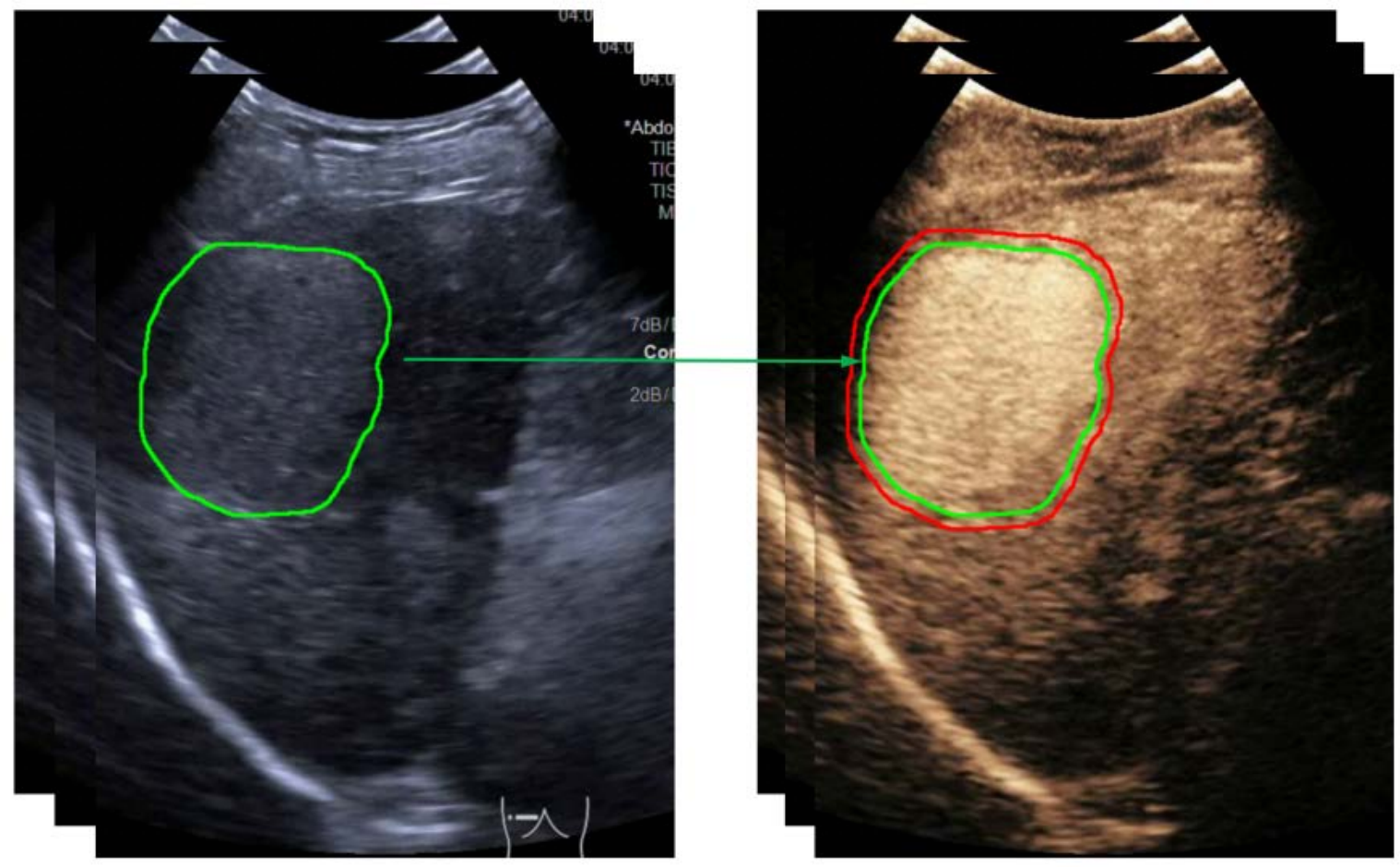

Figure S4. Annotation of DCE-US video sequences. The green contour in the grayscale ultrasound panel indicates the manually delineated tumor boundary (left). This contour was transferred to the spatially corresponding contrast-enhanced panel and expanded to include peritumoral tissue (right).

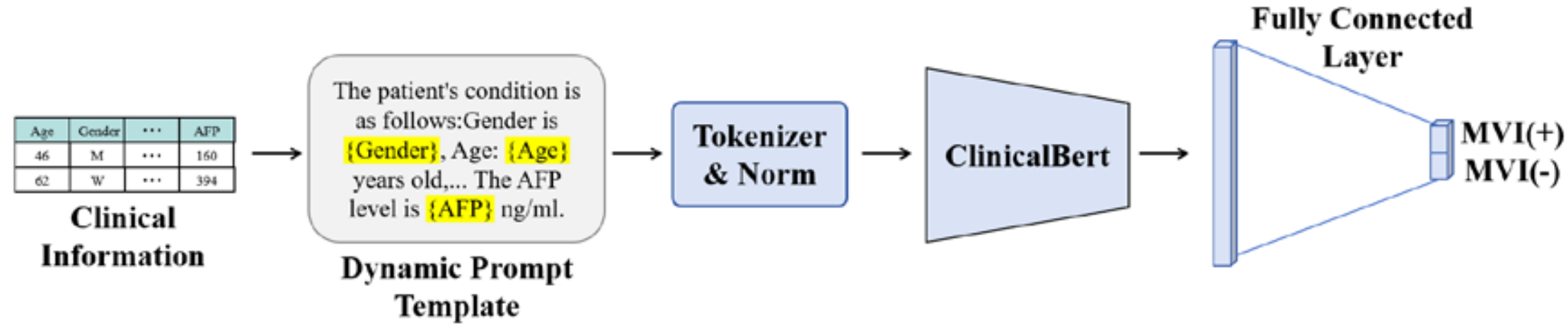

Figure S5. ClinicalBERT-based framework for MVI prediction from clinical information. Clinical variables and manually assessed ultrasound features were converted into standardized text and encoded using ClinicalBERT for MVI prediction.

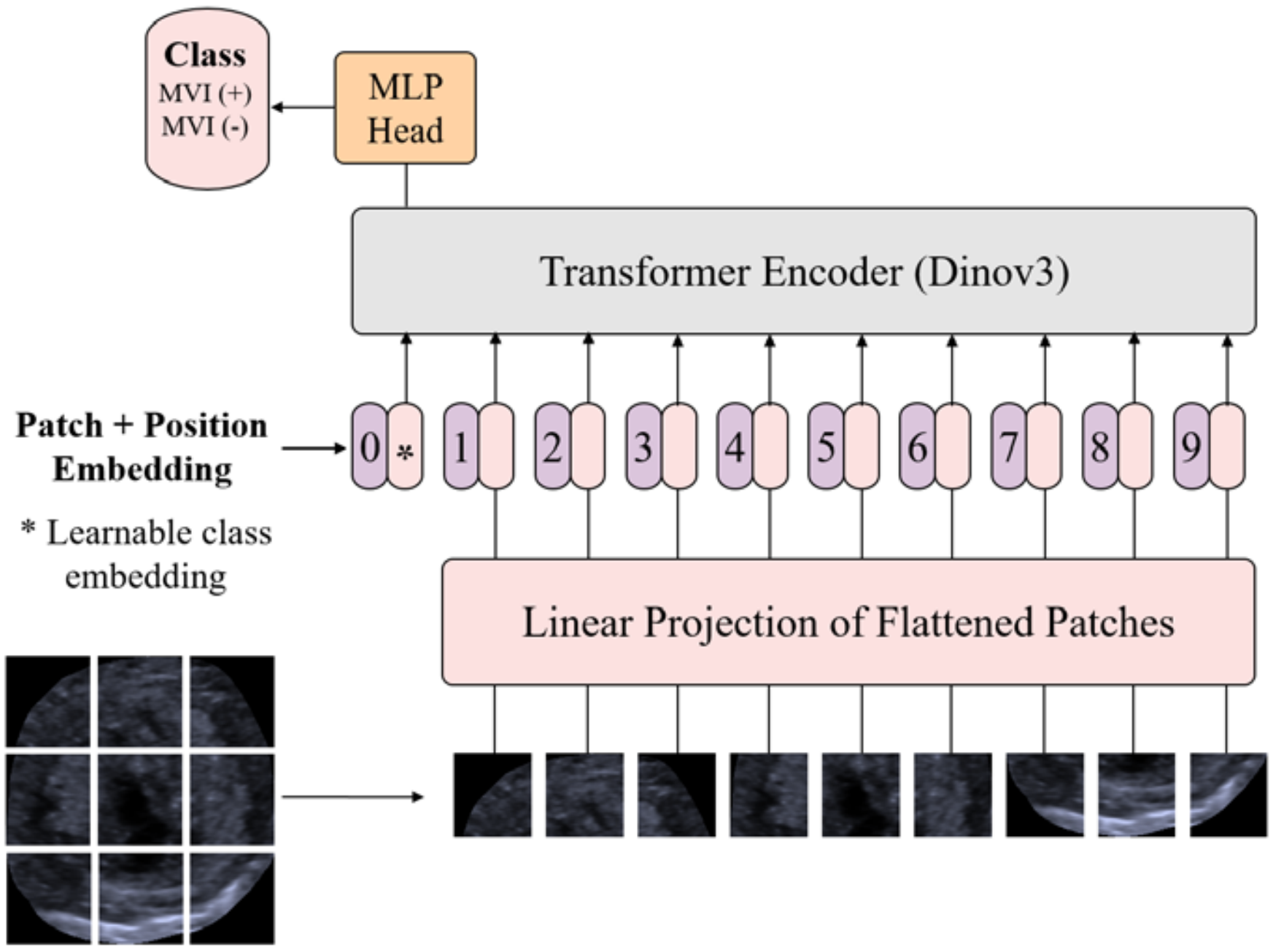

Figure S6. ViT-based framework for MVI prediction from two-dimensional ultrasound images.

## Supplementary Tables

Table S1. Characteristics of the participating centers and cohort composition.

| Cohort | Center No. | Center | City | Administrative level | No. of cases | No. of MVI+ | No. of MVI- |
|---|---|---|---|---|---|---|---|
| Development cohort (7 centers) | 1 | Chinese PLA General Hospital | Beijing | National | 116 | 75 | 41 |
| | 2 | Sun Yat-sen University Cancer Center | Guangzhou | National | 96 | 34 | 62 |
| | 3 | Xiangya Hospital, Central South University | Changsha | National | 83 | 31 | 52 |
| | 4 | Sun Yat-sen Memorial Hospital, Sun Yat-sen university | Guangzhou | National | 52 | 18 | 34 |
| | 5 | Fujian Cancer Hospital | Fuzhou | Provincial | 43 | 25 | 18 |
| | 6 | Harbin Medical University Cancer Hospital | Harbin | Municipal | 20 | 13 | 7 |
| | 7 | Tianjin Third Central Hospital | Tianjin | Municipal | 11 | 9 | 2 |
| External validation cohort (1 center) | 8 | The Third Affiliated Hospital, Sun Yat-sen University | Guangzhou | National | 68 | 25 | 43 |
| Overall | | | | | 489 | 230 | 259 |

Table S2. Distribution of manually assessed ultrasound features across the study cohorts.

| Features | Overall (n = 489) | Development cohort (n = 421) | External validation cohort (n = 68) |
|---|---|---|---|
| Maximum diameter of lesion (cm) | 5.0 (1.2-16.2) | 5.1 (1.2-16.2) | 4.5 (1.2-16.1) |
| Start time of arterial phase enhancement (s)* | 13.0 (5.0-30.0) | 12.8 (5.0-30.0) | 14.4 (6.0-30.0) |
| Time to peak (s)* | 21.0 (10.0-47.0) | 20.7 (11.0-40.0) | 22.5 (10.0-47.0) |
| Rise time* | 8.0 (2.0-19.0) | 7.9 (2.0-17.0) | 8.1 (3.0-19.0) |
| Start time of washout (s)* | 119.9 (25.0-360.0) | 120.3 (25.0-360.0) | 117.6 (38.0-360.0) |
| Tumor infiltration boundary (mm)* | 2.1 (-10.0-27.0) | 2.0 (-10.0-27.0) | 2.4 (-7.0-21.0) |
| Internal blood flow* | | | |
| No | 188 (38.3%) | 153 (36.2%) | 35 (51.5%) |
| Yes | 303 (61.7%) | 270 (63.8%) | 33 (48.5%) |
| Shape of the lesion* | | | |
| Circle | 164 (33.4%) | 162 (38.3%) | 2 (2.9%) |
| Ellipse | 221 (45.0%) | 168 (39.7%) | 53 (77.9%) |
| Foliation | 12 (2.4%) | 11 (2.6%) | 1 (1.5%) |
| Irregular | 67 (13.6%) | 55 (13.0%) | 12 (17.6%) |
| Other | 27 (5.5%) | 27 (6.4%) | 0 |
| Lesion boundary* | | | |
| Clear | 364 (74.1%) | 304 (71.9%) | 60 (88.2%) |
| Ambiguity | 127 (25.9%) | 119 (28.1%) | 8 (11.8%) |
| Tumor echo genicity* | | | |
| Non-hypoechoic | 205 (41.8%) | 186 (44.0%) | 19 (27.9%) |
| Hypoechoic | 286 (58.2%) | 237 (56.0%) | 49 (72.1%) |
| Tumor echo distribution* | | | |
| Uniform | 134 (27.3%) | 92 (21.7%) | 42 (61.8%) |
| Non-uniform | 357 (72.7%) | 331 (78.3%) | 26 (38.2%) |
| Overall enhancement pattern* | | | |
| Uniform | 319 (65.0%) | 274 (64.8%) | 45 (66.2%) |
| Non-uniform | 164 (33.4%) | 141 (33.3%) | 23 (33.8%) |
| No | 8 (1.6%) | 8 (1.9%) | 0 |
| Portal venous phase degree of enhancement* | | | |
| Uniform enhancement | 167 (34.0%) | 143 (33.8%) | 24 (35.3%) |
| Low enhancement | 320 (65.2%) | 276 (65.2%) | 44 (64.7%) |
| High enhancement | 4 (0.8%) | 4 (0.9%) | 0 |
| Late phase degree of enhancement* | | | |
| Uniform enhancement | 25 (5.1%) | 23 (5.4%) | 2 (2.9%) |
| Low enhancement | 466 (94.9%) | 400 (94.6%) | 66 (97.1%) |
| Smooth boundary* | | | |
| Yes | 197 (40.1%) | 170 (40.2%) | 27 (39.7%) |
| No | 294 (59.9%) | 253 (59.8%) | 41 (60.3%) |
| Posterior acoustic enhancement* | | | |
| Normal | 417 (85.5%) | 358 (85.2%) | 59 (86.8%) |
| Enhancement | 69 (14.1%) | 60 (14.3%) | 9 (13.2%) |
| Attenuation | 2 (0.4%) | 2 (0.5%) | 0 |

Table S2 (continued). Distribution of manually assessed ultrasound features across the study cohorts.

| Features | Overall (n = 489) | Development cohort (n = 421) | External validation cohort (n = 68) |
|---|---|---|---|
| Multiple fused nodules* | | | |
| No | 481 (98.4%) | 415 (98.6%) | 66 (97.1%) |
| Yes | 8 (1.6%) | 6 (1.4%) | 2 (2.9%) |
| Hypoechoic halo* | | | |
| No | 346 (70.5%) | 305 (72.1%) | 41 (60.3%) |
| Yes | 145 (29.5%) | 118 (27.9%) | 27 (39.7%) |
| Pseudocapsule* | | | |
| No | 399 (81.3%) | 346 (81.8%) | 53 (77.9%) |
| Yes | 92 (18.7%) | 77 (18.2%) | 15 (22.1%) |
| Necrosis* | | | |
| No | 374 (76.2%) | 328 (77.5%) | 46 (67.6%) |
| Yes | 117 (23.8%) | 95 (22.5%) | 22 (32.4%) |
| Lesion blood flow signal grade | | | |
| 1 | 155 (31.7%) | 133 (31.6%) | 22 (32.4%) |
| 2 | 173 (35.4%) | 146 (34.7%) | 27 (39.7%) |
| 3 | 110 (22.5%) | 95 (22.6%) | 15 (22.1%) |
| 4 | 51 (10.4%) | 47 (11.2%) | 4 (5.9%) |
| Features of the liver background | | | |
| Normal | 40 (8.2%) | 37 (8.8%) | 3 (4.4%) |
| Strong echo | 160 (32.7%) | 154 (36.6%) | 6 (8.8%) |
| Hepatic steatosis | 37 (7.6%) | 29 (6.9%) | 8 (11.8%) |
| Hepatic fibrosis | 186 (38.0%) | 148 (35.2%) | 38 (55.9%) |
| Strong echo + Hepatic steatosis | 4 (0.8%) | 2 (0.5%) | 2 (2.9%) |
| Strong echo + Hepatic fibrosis | 62 (12.7%) | 51 (12.1%) | 11 (16.2%) |

Note. Discrete variables are represented as: number of cases (%) and continuous variables are represented as: median (minimum value - maximum value). * indicates calculation based on the number of tumors.

Table S3. Typical DCE-US acquisition parameters for the ultrasound systems used in this study.

| Manufacturer | System model | SonoVue dose (mL) | Transducer | Frequency range (MHz) | Mechanical index | Frame rate (frames/s) |
|---|---|---|---|---|---|---|
| Philips | EPIQ7 | 1.2-1.6 | C5-1 | 1-5 | 0.06 | 7 |
| Mindray | M9 | 1.8 | C5-1S | 1-5 | 0.096 | 10 |
| | Resona7 | 1.5 | SC5-1U | 1-5 | 0.078 | 11 |
| Siemens | New Sequoia | 1.0-1.5 | 5C1 | 1-5.7 | 0.08-0.10 | 8-24 |
| | S2000 | 1.5-2.4 | 4C1 | 1-4.5 | 0.08-0.10 | 8-12 |
| | Sequoia 512 | 2.0-2.4 | 4C1/4V1 | 1.5-4.5 | 0.18-0.22 | 8-10 |
| HITACHI | ARIETTA 70 | 1.2-2.4 | C251 | 1-5 | 0.09-0.13 | <15 |
| | ARIETTA 850 | 1.2-2.4 | C252 | 1-6 | 0.09-0.13 | <15 |
| Canon | Aplio500/400 | 1.6-2.0 | PVT-375BT | 1.9-6.0 | 0.07-0.08 | 10 |
| Esaote | MyLab8 | 1.6-2.0 | CA541 | 1-8 | 0.08-0.12 | 24 |
| | MyLab Twice | 1.8-2.4 | CA541 | 1-8 | 0.08-0.12 | 24 |
| | MyLab90 | 1.6-2.0 | CA431/CA430E | 1-8 | 0.06-0.08 | 18 |

Table S4. Univariable associations of clinical variables and manually assessed ultrasound features with MVI status in the development cohort.

| Features | P value |
|---|---|
| History of chronic disease | 0.002 |
| Overall enhancement pattern | 0.002 |
| Late phase degree of enhancement | 0.008 |
| Necrosis | 0.009 |
| AFP classification | 0.001 |
| Maximum diameter of lesion | < 0.001 |
| Start time of washout | < 0.001 |
| Rise time | 0.039 |
| AFP level | < 0.001 |
| Tumor infiltration boundary | < 0.001 |

Table S5. External validation performance of clinical information models.

| Method | Model | AUC | ACC (%) | SEN (%) | SPE (%) | F1 (%) |
|---|---|---|---|---|---|---|
| Machine Learning | Logistic regression | 0.6491±0.0376 | 60.88±6.46 | 60.74±8.11 | 60.98±10.63 | 55.23±5.95 |
| | XGBoost | 0.6180±0.0471 | 59.12±2.42 | 51.85±12.83 | 63.90±9.51 | 49.56±6.88 |
| | SVM | 0.6392±0.0253 | 61.76±2.75 | 54.07±7.22 | 66.83±5.34 | 52.76±4.37 |
| | Random Forest | 0.6172±0.0409 | 60.00±3.19 | 57.04±8.92 | 61.95±4.43 | 52.89±5.58 |
| Deep Learning | Multilayer perceptron | 0.5991±0.0739 | 54.41±8.76 | 56.30±9.59 | 53.17±19.99 | 49.55±2.67 |
| | Graph convolutional network | 0.5693±0.1089 | 56.47±2.67 | 45.93±33.19 | 63.41±20.84 | 39.17±25.33 |
| | ClinicalBERT | 0.6715±0.0156 | 59.41±6.94 | 60.00±17.45 | 59.02±21.50 | 53.50±5.14 |

Table S6. External validation performance of BUS and CDFI image models.

| Model | Pretraining | Modality | AUC | ACC (%) | SEN (%) | SPE (%) | F1 (%) |
|---|---|---|---|---|---|---|---|
| ResNet-18 | ImageNet | BUS | 0.5650± 0.0918 | 50.88± 10.94 | 62.40± 7.80 | 44.19± 19.60 | 48.67± 5.57 |
| | | CDFI | 0.6400± 0.0626 | 58.82± 5.30 | 51.20± 16.59 | 63.26± 11.08 | 47.01± 9.28 |
| Swin-T | ImageNet | BUS | 0.5775± 0.0335 | 53.53± 11.70 | 55.20± 27.04 | 52.56± 33.77 | 44.14± 10.43 |
| | | CDFI | 0.6033± 0.0756 | 56.76± 4.94 | 52.80± 3.35 | 59.07± 8.48 | 47.41± 2.83 |
| ViT-B | ImageNet | BUS | 0.5708± 0.0366 | 54.12± 9.44 | 61.60± 25.39 | 49.77± 28.70 | 47.94± 9.08 |
| | | CDFI | 0.6247± 0.0702 | 59.12± 4.34 | 59.20± 20.67 | 59.07± 16.08 | 50.24± 9.67 |
| ViT-B | MAE | BUS | 0.5715± 0.0144 | 45.29± 2.63 | 73.60± 17.34 | 28.84± 10.61 | 49.22± 5.90 |
| | | CDFI | 0.5901± 0.0364 | 50.88± 4.24 | 68.00± 10.20 | 40.93± 9.10 | 50.27± 4.54 |
| ViT-B | DINOv3 | BUS | 0.6087± 0.0417 | 58.24± 5.75 | 59.20± 21.05 | 57.67± 19.83 | 49.93± 7.00 |
| | | CDFI | 0.6435± 0.0344 | 57.94± 7.25 | 65.60± 19.51 | 53.49± 18.09 | 52.67± 7.80 |

Table S7. Performance comparison of DCE-US frame sampling strategies in the external validation cohort. Values represent results obtained using the first cross-validation fold.

| Frame sampling strategy | AUC | ACC (%) | SEN (%) | SPE (%) | F1 (%) |
|---|---|---|---|---|---|
| Random | 0.7963 | 76.47 | 68.00 | 81.40 | 68.00 |
| Uniform | 0.8326 | 77.94 | 88.00 | 72.09 | 74.58 |
| Pixel difference | 0.8521 | 73.53 | 76.00 | 72.09 | 67.86 |

Table S8. External validation performance of different fusion strategies for BUS, CDFI, and DCE-US representations.

| **Fusion strategy** | **AUC** | **ACC (%)** | **SEN (%)** | **SPE (%)** | **F1 (%)** |
|---|---|---|---|---|---|
| Addition | 0.8268± 0.0328 | 70.00± 5.85 | 80.80± 17.06 | 63.72± 17.76 | 66.19± 3.90 |
| Average | 0.7801± 0.0800 | 71.18± 7.75 | 68.80± 16.10 | 72.56± 13.50 | 63.54± 9.24 |
| Concatenation | 0.8305± 0.0197 | 75.88± 3.05 | 68.00± 14.97 | 80.47± 8.48 | 66.83± 7.28 |
| Bilinear Fusion | 0.8266± 0.0302 | 76.76± 6.01 | 67.20± 15.59 | 82.33± 17.68 | 67.98± 3.59 |
| Attention Bottleneck | 0.8655± 0.0244 | 79.41± 3.75 | 72.80± 10.35 | 83.26± 9.78 | 72.20± 3.58 |
| Transformer (1 layer) | 0.8681± 0.0185 | 71.76± 6.36 | 83.20± 7.69 | 65.12± 13.56 | 68.70± 4.01 |
| Transformer (2 layers) | 0.8860± 0.0121 | 79.41± 1.80 | 72.80± 10.73 | 83.26± 8.61 | 72.05± 2.26 |
| Transformer (3 layers) | 0.8759± 0.0174 | 79.41± 1.47 | 68.80± 11.10 | 85.58± 6.86 | 70.80± 3.43 |

Table S9. Performance of ultrasound physicians for MVI assessment in the external validation cohort.

| **Ultrasound physicians** | **ACC (%)** | **SEN (%)** | **SPE (%)** | **F1 (%)** | **kappa** |
|---|---|---|---|---|---|
| 1 | 54.41 | 64.00 | 48.84 | 50.79 | 0.4202 |
| 2 | 44.62 | 64.00 | 32.50 | 47.06 | |